\documentclass{article}
\usepackage{ijcai24}

\usepackage{times}
\usepackage{soul}
\usepackage{url}
\usepackage[hidelinks]{hyperref}
\usepackage[utf8]{inputenc}
\usepackage[small]{caption}
\usepackage{graphicx}
\usepackage{amsthm}
\usepackage{amsmath}
\usepackage{booktabs}
\usepackage{algorithm}
\usepackage{algorithmic}
\usepackage[switch]{lineno}

\usepackage{subcaption}
\usepackage{amsfonts}       
\usepackage{nicefrac}       
\usepackage{microtype}      
\usepackage{enumitem}       
\setlist[enumerate]{leftmargin=*}       
\usepackage{cleveref}       
\usepackage{lipsum}         
\usepackage{doi}
\usepackage[dvipsnames]{xcolor}
\usepackage{array,multirow,graphicx}

\DeclareMathOperator*{\argmin}{\arg\!\min}
\DeclareMathOperator{\Var}{\text{Var}}

\newcommand{\citet}[1]{\citeauthor{#1} [\citeyear{#1}]}
\newcommand{\citep}[1]{\cite{#1}}
\newcommand{\cited}[2]{[\citeauthor{#1}, \citeyear{#1}; \citeauthor{#2}, \citeyear{#2}]}
\newcommand{\termdefn}[1]{\emph{#1}}

\newcommand*{\img}[1]{%
    \raisebox{-.2\baselineskip}{%
        \includegraphics[
        height=\baselineskip,
        width=\baselineskip,
        keepaspectratio,
        ]{#1}%
    }%
}

\newtheorem{example}{Example}
\newtheorem{theorem}{Theorem}
\newtheorem{assumption}{Assumption}

\title{Robust Losses for Decision-Focused Learning}

\author{
    Paper 1601
}

\author{
Noah Schutte$^1$
\and
Krzysztof Postek$^2$\and
Neil Yorke-Smith$^1$
\affiliations
$^1$Delft University of Technology\\
$^2$Independent Researcher
\emails
\{n.j.schutte, n.yorke-smith\}@tudelft.nl,
krzysztof.postek@gmail.com
}

\begin{document}

\maketitle

\begin{abstract}
Optimization models used to make discrete decisions often contain uncertain parameters that are context-dependent and estimated through prediction. To account for the quality of the decision made based on the prediction, decision-focused learning (end-to-end predict-then-optimize) aims at training the predictive model to minimize regret, i.e., the loss incurred by making a suboptimal decision. Despite the challenge of the gradient of this loss w.r.t. the predictive model parameters being zero almost everywhere for optimization problems with a linear objective, effective gradient-based learning approaches have been proposed to minimize the expected loss, using the empirical loss as a surrogate. However, empirical regret can be an ineffective surrogate because empirical optimal decisions can vary substantially from expected optimal decisions. To understand the impact of this deficiency, we evaluate the effect of aleatoric and epistemic uncertainty on the accuracy of empirical regret as a surrogate. Next, we propose three novel loss functions that approximate expected regret more robustly. Experimental results show that training two state-of-the-art decision-focused learning approaches using robust regret losses improves test--sample empirical regret in general while keeping computational time equivalent relative to the number of training epochs.
\end{abstract}


\section{Introduction}
\label{sec:intro}

Real-world optimization problems, often formulated and solved as mixed-integer linear problems (MIPs) -- such as shortest path problems or machine scheduling problems -- involve parameters whose value is not known exactly. It is natural to use data to predict the uncertain parameters' values based on contextual information.

In predictive (regression) problems the goal is to make the most accurate prediction possible in the sense of prediction error. However, since there remains uncertainty around the predictions, and their purpose is to more accurately solve a downstream optimization problem, such a measure of prediction accuracy is of little relevance: the quality of the resulting decisions is what is important.  This is the main premise of \termdefn{decision-focused learning} (DFL) (also named predict+optimize or smart/end-to-end predict-then-optimize), and contrasts with \termdefn{prediction-focused learning} (PFL) which is focused on prediction accuracy. DFL was first pioneered specifically for portfolio selection by \citet{Bengio1997}, and more generally introduced by \citet{Elmachtoub2022}.

Learning based on decision errors obtained through solving a MIP faces difficulty, since the gradient of the decisions relative to the predicted parameters is zero almost everywhere and otherwise undefined \citep{Vlastelica2019}.  Because of this, approximation methods have been introduced and shown to be more effective than PFL \cited{Elmachtoub2022}{Berthet2020}.  The loss used in these methods is based on \termdefn{empirical regret}, i.e., the loss incurred from not making the optimal decision for a given, empirical scenario. However, we observe that minimizing empirical regret deviates from minimizing expected regret as it leads to a form of overfitting on the empirical samples.

Recognizing this shortcoming in the most commonly used loss function in DFL, this paper develops three contributions:
\begin{enumerate}
    \item We examine the usage of empirical regret from an uncertainty perspective, showing how empirical regret can lead to poor generalization and biased learning towards uncertain parameters with high variance. 
    \item We propose three different robust loss functions that simultaneously (1) improve conditional mean estimation, (2) are robust against errors in the mean estimation, and (3) do not increase computational expense relative to the number of training epochs.
    \item We study 
    two state-of-the-art gradient approximation DFL methods using the proposed robust losses compared to them 
    using empirical regret.  
    On 
    three experimental problems, average test--sample regret is improved without additional overhead. 
\end{enumerate}

\section{Problem Formulation}
\label{sec:problem}

\begin{figure*}[t!]
    \centering
    \includegraphics[width=\textwidth]{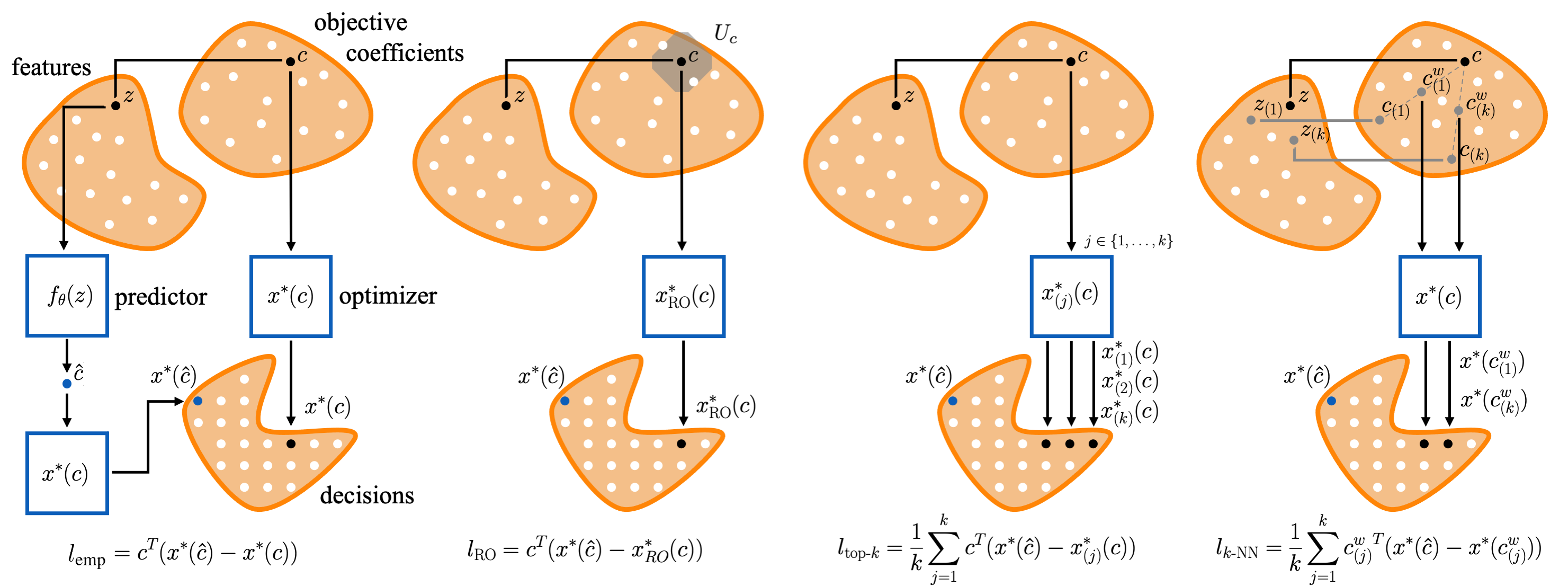} 
    \caption{Visualization of the empirical regret loss in DFL (left) and the robust losses in comparison (predictive pipeline is equal).  
    The robust losses are constructed through using an optimization model (optimizer) that is robust against the mean estimation error and/or using a different mean estimator than empirically observed $c$.
    From left to right: empirical regret ($l_\text{emp}$), RO loss ($l_\text{RO}$), top-$k$ loss ($l_\text{top-$k$}$), $k$-NN loss ($l_\text{$k$-NN}$).
    }
    \label{fig:robloss}
\end{figure*}

In this work we adopt the problem setting as introduced in \citet{Elmachtoub2022}.  Observing some contextual information in the form of feature values $z \in \mathbb{R}^m$, the goal of the decision maker is to solve the following stochastic optimization problem with linear objective:
\begin{align} \label{eq:sto}
    \min_{x \in X} \mathbb{E}_{c \sim \mathcal{C}_z}[c^T x|z],  
\end{align}
where $X$ denotes the set of feasible decisions and $\mathcal{C}_z$ is the conditional distribution of uncertain objective coefficients $c \in \mathbb{R}^n$ given feature values $z$. Since $\mathcal{C}_z$ is unknown, training a 
parametric predictor $f_\theta(z)$ can assist in picking good decisions. Ideally we find a predictor s.t.\@ for each $z$ we make a decision that has an equal objective to 
\eqref{eq:sto}, i.e.,
\vspace{-1mm}
\begin{align*}
     \mathbb{E}_{c \sim \mathcal{C}_z}[c|z]^T \left[ \argmin_{x \in X} f_\theta(z)^T x \right] = \min_{x \in X} \mathbb{E}_{c \sim \mathcal{C}_z}[c^T x|z]. 
\end{align*}
 When the prediction problem is seen as separate from the optimization problem (PFL), regression is performed with coefficients $c$ as responses to features $z$ by minimizing prediction error.  However, since the decision maker's goal is not to make accurate predictions but to make optimal decisions, it is preferable to minimize decision error, which is the main premise of DFL.  To attain this goal, \citet{Elmachtoub2022} introduce a loss function based on the notion of \textit{regret}, i.e., the loss incurred by not making the optimal decision given $z$:
\vspace{-1mm}
\begin{equation} \label{eq:emploss}
    l_{\text{emp}}(\hat{c}, c) = c^T x^*(\hat{c}) - c^T x^*(c), 
\end{equation}
where $\hat{c} = f_\theta(z)$, and $x^*(c) = \argmin_{x \in X} c^T x$ is the optimal decision when $c$ are the objective coefficients.
We will refer to this as \termdefn{empirical regret}, as it is the regret in an empirical realization of $(z, c)$. This loss is a natural choice, for when $(z, c)$ is realized the loss is minimal when the empirical optimal decision is made.  Figure \ref{fig:robloss} visualizes this loss and the robust losses that will be introduced in Section \ref{sec:losses}.

Due to the effectiveness of (stochastic) gradient descent in training predictive models, much research on DFL concentrates on finding effective gradient approximations. Approximations are necessary when the optimization problem is defined as in \eqref{eq:sto}, as the linearity of the uncertain parameters in the objective function makes the true gradient zero almost everywhere and otherwise undefined. Gradient descent is based on the premise that minimizing empirical losses leads to minimizing the expected loss. The empirical loss is used as a surrogate, while the main goal is to find predictive parameters such that for every feature realization $z$ we have a minimal expected regret loss defined as: 
\vspace{-1mm}
\begin{equation} \label{eq:loss2}
    l_{\mathbb{E}}(\hat{c}, \mathcal{C}_z) = \mathbb{E}_{c \sim \mathcal{C}_z}[c|z]^T x^*(\hat{c}) - \min_{x \in X} \mathbb{E}_{c \sim \mathcal{C}_z}[c^T x|z].
\end{equation}
Directly minimizing this loss is often impossible since in practice data is only available in the form of realized value pairs $D = \{(z_i, c_i) : i \in \{1, \dots, t\}\}$. The empirical loss is a valid alternative, as the set of minimizers of \eqref{eq:loss2} is equal to the set of minimizers of the expectation of \eqref{eq:emploss}, i.e.:
\vspace{-1mm}
\begin{equation*} \label{eq:minimizers}
    \min_\theta \{l_{\mathbb{E}}(f_\theta(z), \mathcal{C}_z)\} = \min_\theta \{ \mathbb{E}_{c \sim \mathcal{C}_z}[l_{\text{emp}}(f_\theta(z), c)] \},
\end{equation*}
and therefore the assumption in the literature until now is that representative data should be sufficient to learn good predictive models through empirical regret. In this paper we challenge this assumption and show that when there is significant uncertainty empirical regret can be an ineffective surrogate. We propose three alternative losses that are more robust.

We begin by showing what issues can arise when training using empirical regret. The main intuition is the following: \textit{When using empirical regret, training is biased towards empirical optimal decisions that are not necessarily expected optimal decisions}. This is due to a form of overfitting, i.e., uncertainty causing a mismatch between training and test data, and the non-smoothness of a discrete optimization problem, which can cause significant changes in optimal decisions by only small perturbations. To analyze the impact of uncertainty more formally, we make the distinction between \termdefn{epistemic} and \termdefn{aleatoric} \citep{Kiureghian2009}.  Epistemic uncertainty is uncertainty that occurs due to a lack of knowledge, which in practice often equates to not having enough representative data available. Aleatoric uncertainty is the uncertainty inherent to some underlying process and is therefore irreducible. In our DFL setting, we assume there exists a distribution $\mathcal{C}_z$ for every $z$ that contains purely aleatoric uncertainty, but since we are limited by finite data the distributions $\mathcal{C}_z$ are unknown to us -- the epistemic uncertainty. 

\begin{figure*}[tb]
    \centering
    \begin{subfigure}[t]{0.35\textwidth}
        \centering
        \includegraphics{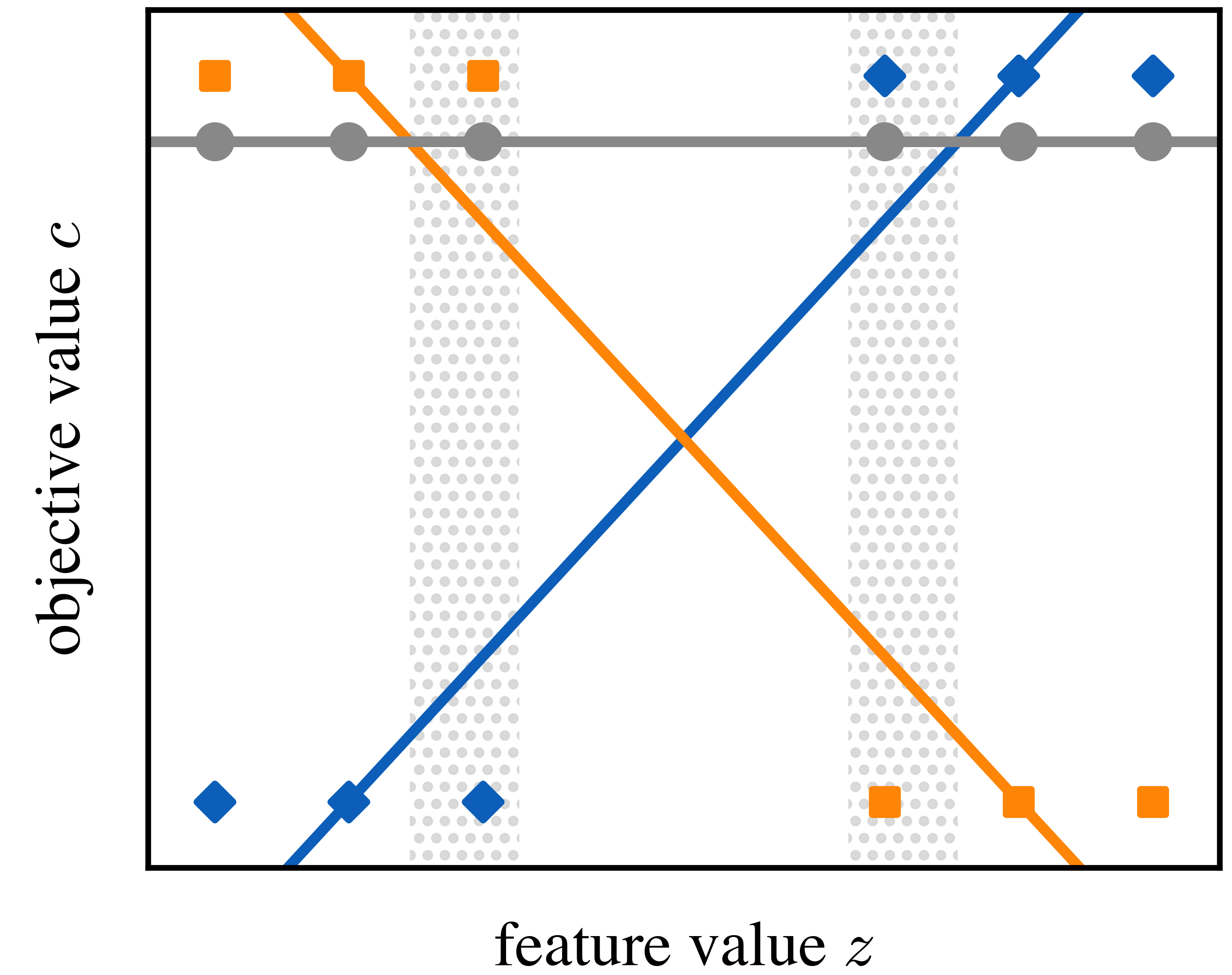}
        \caption{PFL -- mean squared error}
        \label{fig:exa}
    \end{subfigure}
    \begin{subfigure}[t]{0.32\textwidth}
        \centering
        \includegraphics{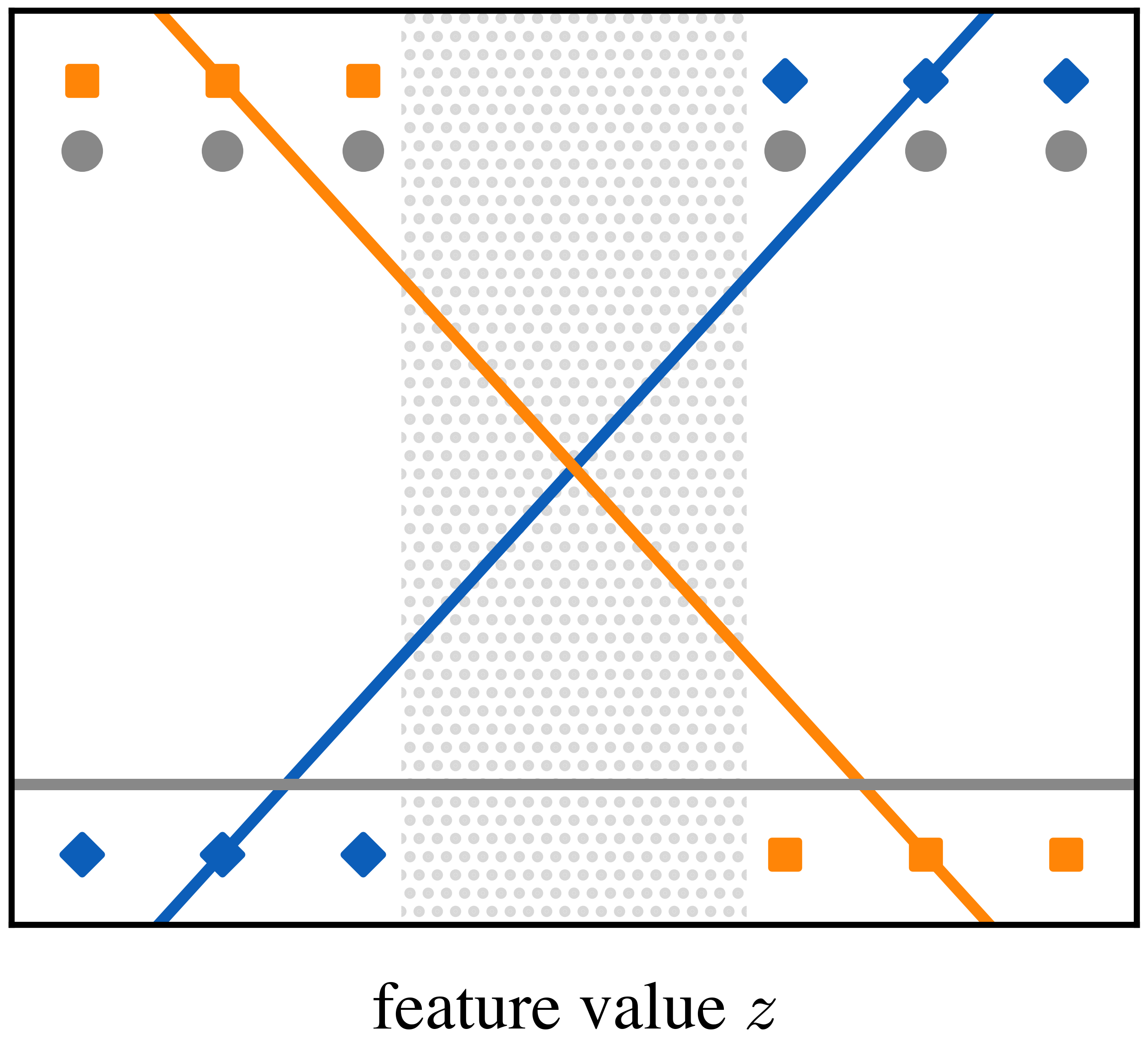}
        \caption{DFL -- empirical regret}
        \label{fig:exb}
    \end{subfigure}
        \begin{subfigure}[t]{0.32\textwidth}
        \centering
        \includegraphics{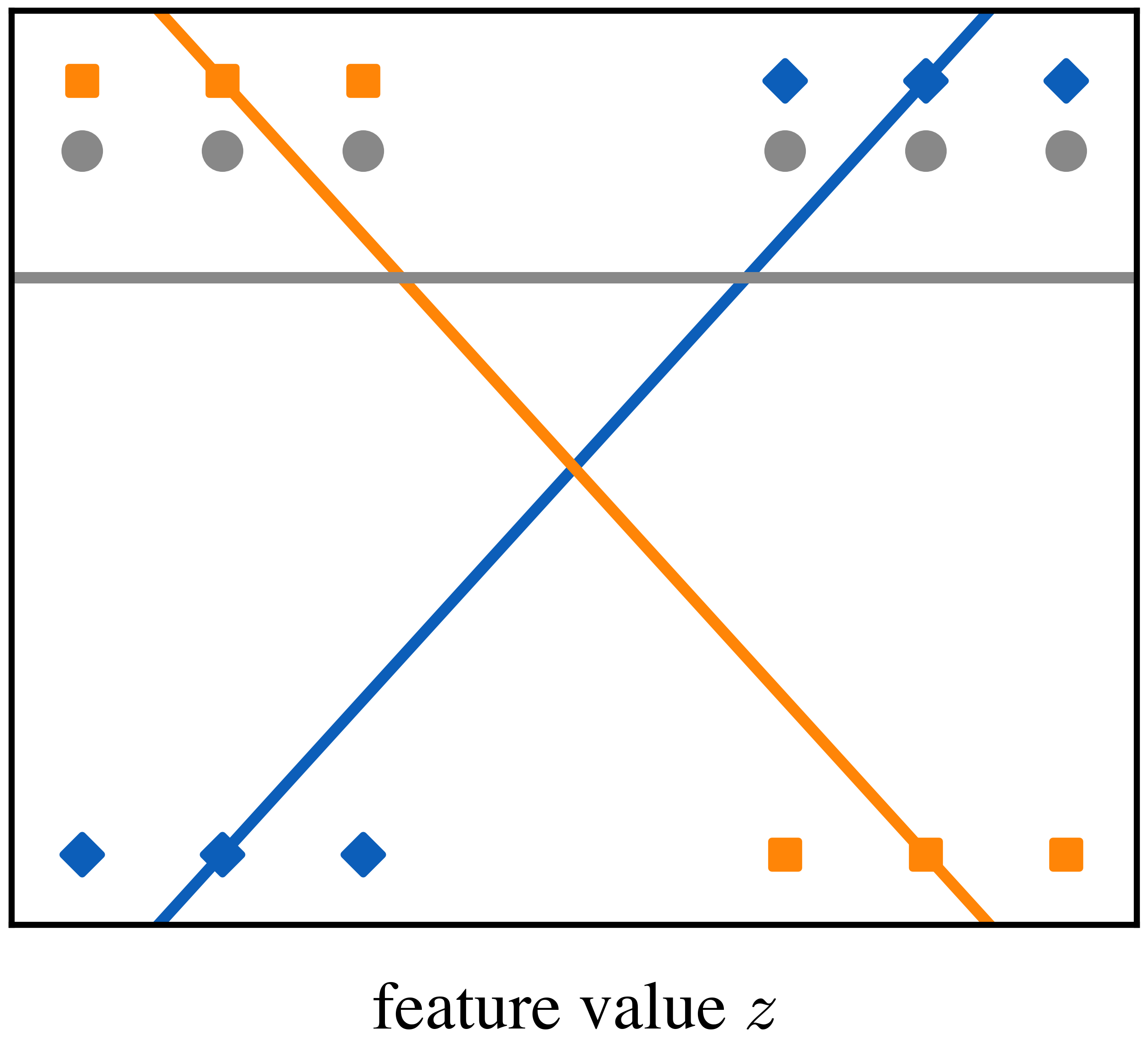}
        \caption{DFL -- robust regret}
        \label{fig:exc}
    \end{subfigure}
    \vspace{-2mm}
    \caption{Observed coefficients $c$ (profit) per feature value $z$ (temperature) and optimal linear predictors $\hat{c} = f_{\theta}^*(z)$ according to different loss functions. The example problem is a coffee stand owner deciding what treat to make from their daily batch of chocolate. There are 3 possible weather related decisions: chocolate ice cream \img{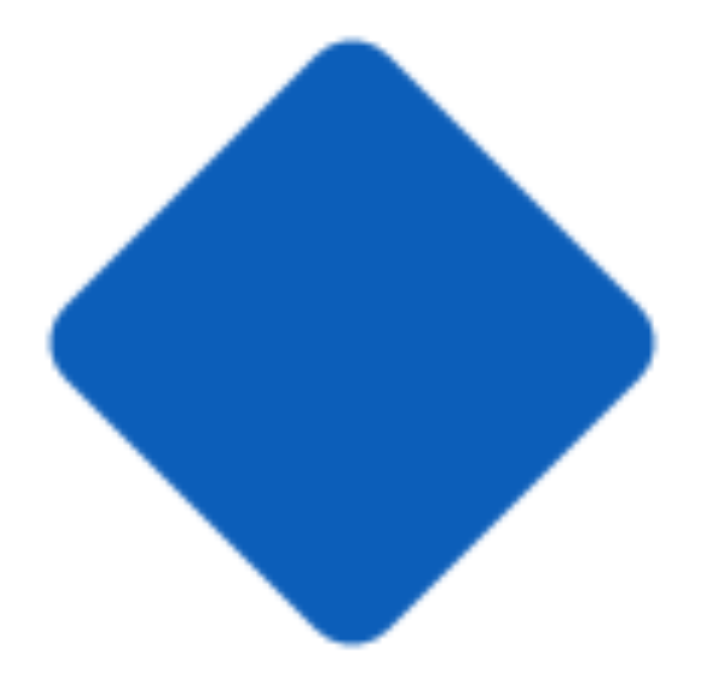}, chocolate cookies \img{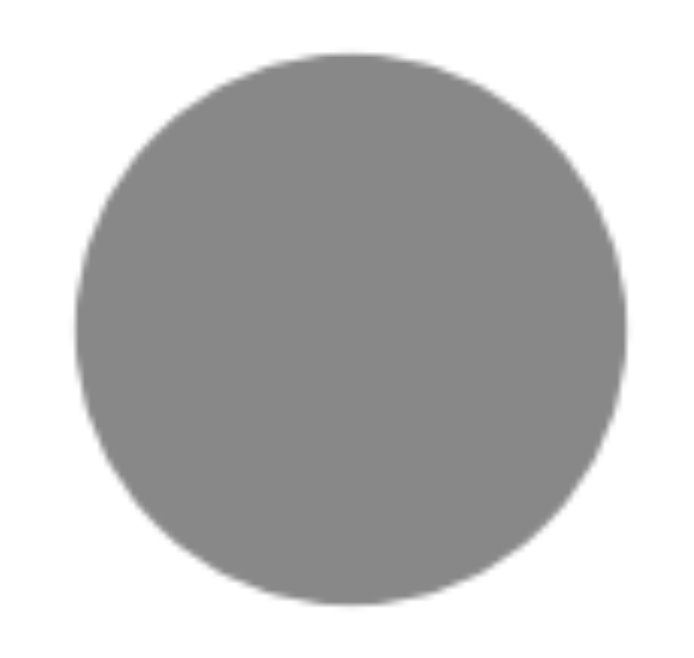} or hot chocolate \img{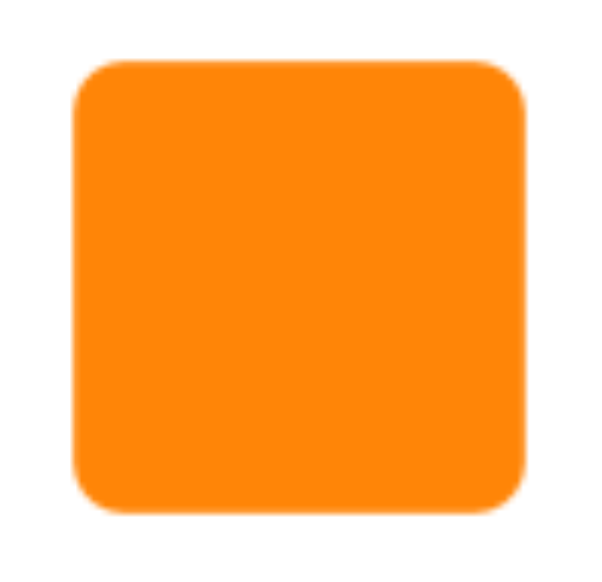}. In short: $\max_{x} c^T x \text{ s.t. } x_i \in \{0, 1\}, \sum_i x_i = 1$. The linear predictors are the lines with corresponding colour and shade. The shaded area  \img{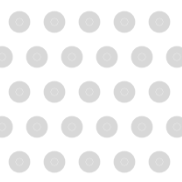} is considered to be sub-optimal. (It is determined assuming the expected value of $c$ is equal to a linear interpolation between the observed points closest to the middle.) 
}
    \label{fig:ex}
\end{figure*}

\subsection{Impact of Epistemic Uncertainty}
Epistemic uncertainty makes generalization, i.e., adapting to unseen data, challenging as there is no complete knowledge, hence making it impossible to learn the exact underlying process. This general challenge for predictive problems is exemplified in DFL due to the focus on optimal decisions.  When using PFL, the predictive model is trained to predict all coefficients as accurately as possible.  When using DFL with empirical regret, the focus is on predicting the objective coefficients related to optimal decisions, i.e., objective coefficients that directly impact the empirically optimal decisions' objectives. This is because empirical regret \eqref{eq:emploss} is based on the obtained objective given the optimal decision according to prediction $\hat{c}$. This means that if we would like to move towards the empirical optimal loss, i.e., $\hat{c}$ such that $c^T x^*(\hat{c})=c^T x^*(c)$, only objective coefficients that relate to the empirical optimal objective value based on prediction $\hat{c}$ are relevant. Other objective coefficients are disregarded, which can lead to poor generalization.

Figure~\ref{fig:ex} gives an illustrative example in which PFL using the mean squared error and DFL using empirical regret both lead to likely sub-optimal predictors. Note that in Figure~\ref{fig:exb}, the linear predictors are empirically optimal as long as square (\img{square}) is predicted with low values of $z$ and diamond (\img{diamond}) with high values of $z$, which makes the linear predictor of circle (\img{circle}) optimal as long as it does not predict the highest value for any of the realized points $(c, z)$.  This means the predictors denoted in Figure~\ref{fig:exc} are also optimal according to empirical regret, but it has such a wide range of parameter values that are considered optimal it is likely the linear predictor of circle (\img{circle}) predicts low objective values.

\subsection{Impact of Aleatoric Uncertainty}
In general, a lack of enough representative data leads to epistemic uncertainty.  However, even when there is plenty of representative data and only aleatoric uncertainty, we observe that using the empirical loss biases learning towards objective coefficients that have high variance, with variance being a measure of aleatoric uncertainty. This is because the predictive model parameters that affect high variance objective coefficient predictions $\hat{c}$ are more likely to be updated during training than those that affect low variance objective coefficient predictions $\hat{c}$. This can be problematic because it can make predictions of low variance objective coefficients less accurately (also with respect to decision quality), while it is these objective coefficients that relate to optimal decisions with a less uncertain objective value. 

To provide intuition behind this, we again note that empirical regret \eqref{eq:emploss} is based on the obtained objective given the optimal decision according to prediction $\hat{c}$. Now we extend previous reasoning to a probabilistic perspective: If certain coefficients are \textit{more often} related to the empirical optimal objective value, they are more often updated during training, i.e., \termdefn{biased learning} occurs. We provide an example of a simple optimization problem to show that the variance of objective coefficient can affect this. This is specific to DFL compared to regression due to the downstream optimization problem, as $\Var (\min c ) \neq \min \Var(c)$ for some random variable $c$.

\begin{example}
Consider optimization problem \eqref{eq:sto}, where decision set $X$ is finite and each decision has its own objective coefficient, i.e., $X = \{x : x \in \{0, 1\}^n, \sum_{i=1}^n x_i = 1 \}$, where we recall that $c \in \mathbb{R}^n$. Assume that all $c_i$ are independently distributed with $\mathbb{E}_{c \sim \mathcal{C}_z}[c_i|z] = 0$ but different variances. In expectation, all decisions have equal objective value $0$. However, empirically we observe the decision with the lowest realized coefficient as optimal. If the coefficient of decision variable $x_i$ has $\Var(c_i) = 0$,  it is only optimal if all other coefficients are realized greater or equal to $0$.  If the coefficient of $x_i$ has high variance, it is more likely for $x_i$ to be optimal as its coefficient is more likely to realize the lowest value. Considering a dataset with realized values, decisions with higher variance coefficients are more often empirically optimal assuming similarly shaped distributions.
\end{example}

\noindent
A formal proof of this example 
is provided in the Appendix.


\section{Robust Losses}
\label{sec:losses}

We introduce novel DFL losses with the goal to mitigate the issues of poor generalization and biased learning that can arise when using empirical regret as a loss. We show that these issues can be mitigated by finding an estimator of $\mathbb{E}_{c \sim \mathcal{C}_z}[c|z]$ with lower variance than $c$ and making the loss robust against the estimation error.  Based on this we define three novel loss functions.  
First, we note that the predictor $f(z) = c$ would lead to an empirical regret loss of $0$ that is unattainable due to $\mathcal{C}_z$ being the underlying distribution given $z$. A perfect predictor would be $f(z) = \mathbb{E}_{c \sim \mathcal{C}_z}[c|z]$, which is because of the assumption of the uncertain parameters $c$ being linear in the objective function, as
\begin{equation*} 
    \min_{x \in X} \mathbb{E}_{c \sim \mathcal{C}_z}[c^T x|z] = \min_{x \in X} \mathbb{E}_{c \sim \mathcal{C}_z}[c|z]^Tx
\end{equation*}
leads to an expected regret loss of $0$. This is important as it shows that a deterministic predictor can be expressive enough and we do not need to be able to predict the whole distribution $\mathcal{C}_z$ to have a strong predictor. This also means that we can still consider the deterministic problem $x^*(f(z))$ as the problem the decision maker solves when observing feature values $z$ in practice. This consideration is important in designing the robust losses, as this means that during training we can still evaluate the loss by solving a deterministic problem without losing performance. Since DFL using gradient descent is already relatively expensive due to frequently having to solve an optimization problem, it is important computation time is not increased further \citep{Mulamba2021}. 

To find a good surrogate for the expected regret loss, we first look at the empirical regret as surrogate. When we optimize using empirical regret, effectively the realized $c$ are used as an estimate of $\mathbb{E}_{c \sim \mathcal{C}_z}[c|z]$ when considering the expected regret as the true regret. Since the realized $c$ is a sample realization from $\mathcal{C}_z$, the expected estimation error is equal to the variance of $\mathcal{C}_z$. This means that the empirical optimal decision $x^*(c)$ can be significantly different from the expected optimal decision $x^*(\mathbb{E}_{c \sim \mathcal{C}_z}[c|z])$, which leads us to two main principles in designing the robust losses:
\begin{enumerate}
    \item Determine an estimator of $\mathbb{E}_{c \sim \mathcal{C}_z}[c|z]$ that has lower variance than $c$. 
    \item Determine close-to optimal decisions that are robust against the estimation error of $\mathbb{E}_{c \sim \mathcal{C}_z}[c|z]$ as an alternative to $x^*(c)$. 
\end{enumerate}

We introduce two losses based on the second principle, followed by a loss based on the first principle that naturally extends to a generalization that also includes the second principle.  Figure~\ref{fig:robloss} visualizes the three proposed losses compared to empirical regret.

\subsection{Robust Optimization (RO) Loss}
Assuming $c$ is an erroneous estimation of $\mathbb{E}_{c \sim \mathcal{C}_z}[c|z]$, we consider empirical optimal decision $x^*(c)$ as an erroneous optimal decision. Instead of using $x^*(c)$, we can use the decision that minimizes the worst-case value of this error given some assumptions, i.e., 
robust optimal decision
\begin{align} \label{eq:ro}
    x_{\text{RO}}^*(c) = \min_{x \in X} \max_{c \in U_c} c^T x,
\end{align}
where $U_c$ is an \textit{uncertainty set}, i.e., a set that specifies the possible estimation error of $c$. This formulation is commonly used in Robust Optimization (RO), where the goal is to find decisions that are optimal given the worst case in some specified uncertainty set \citep{Bertsimas2011}.  

Since in RO the retrieved optimal decision is robust against the uncertainty specified by the uncertainty set, a proper specification of the latter is important. An uncertainty set that is too small makes the optimal decision not robust against the actual uncertainty, while an uncertainty set that is too large leads to conservative decisions that could be far from expected optimal. Further, depending on the uncertainty set the optimization problem \eqref{eq:ro} has a tractable reformulation. We propose the RO loss defined as follows:
\begin{equation*}
    l_{\text{RO}}(\hat{c}, c) = c^T  (x^*(\hat{c}) - x^*_{RO}(c)). 
\end{equation*}
As a choice of uncertainty set we propose the \termdefn{budget uncertainty set}, which is a special case of a polyhedron uncertainty set \citep{BertsimasHertog2022}. This set is practical as it is not too conservative while it allows for a tractable reformulation of linear optimization problems, i.e., if the optimization problem has a linear objective and only linear constraints specifying the decision set $X$ the reformulation does not increase complexity. Our setting is not limited to this model class, but having a linear tractable reformulation is helpful as it does not change the class of the problem: the budget uncertainty set reformulation of the MIPs we consider is a MIP.  Since the robust formulation adds constraints and auxiliary decision variables, the problem can become harder (albeit also easier) to solve. However this is highly dependent on the problem and is not straightforward to quantify.   

The budget uncertainty set allows us to model the estimation error of $c$ as percentage deviation $\zeta \in \mathbb{R}^n$. We limit the individual coefficient percentage deviation and the total percentage deviation by $\rho$ and $\Gamma$ respectively, giving 
\begin{align*}
    U_c = \{ c \circ (1 + \zeta): ||\zeta||_\infty \leq \rho, ||\zeta||_1 \leq \Gamma \},
\end{align*}
where $\circ$ denotes the Hammard product. Depending on knowledge of the optimization problem and/or the uncertainty around $c$ the uncertainty set can be adjusted. Our goal here is to show a proof of concept, leaving for the future a full exploration of specifying uncertainty sets in a DFL setting.

\subsection{Best \texorpdfstring{$k$}{k} Decisions (Top-\texorpdfstring{$k$}{k}) Loss}
An alternative to finding a decision that is robust against erroneous estimations is to consider multiple decisions that are close-to optimal.  For instance, suppose that in the example in Figure~\ref{fig:ex} the second-best decision is also considered as optimal, then its coefficients would not be disregarded. Looking at multiple close-to optimal decisions can lead to a good area in the decision space, compared to a single optimal decision that might lay in a narrow global optimum. Moreover, high variance of $c$ as an estimate of $\mathbb{E}_{c \sim \mathcal{C}_z}[c|z]$ can lead to significantly different decisions, while considering multiple close-to optimal decisions can lead to certain decisions being found more often and therefore a more stable signal. 

The process of finding multiple quality decisions is not complicated, as any solving method can be used subsequently while excluding already found decisions. Some methods even record suboptimal decisions and therefore a single solve could suffice. This makes the loss we propose viable, as we propose a loss that evaluates regret against the best $k$ decisions (top-$k$):
\vspace{-1mm}
\begin{equation*}
    l_{\text{top-}k}(\hat{c}, c) = \frac{1}{k}\sum_{j = 1}^k c^T  (x^*(\hat{c}) - x^*_{(j)}(c)),
\end{equation*}
where $x^*_{(j)}(c) := \argmin_{x\in X \backslash \{x^*_{(1)}(c), \dots, x^*_{(j-1)}(c) \} } \{ c^T x \}$. 

Alternatively one could refrain from specifying the number of decisions that is considered as good enough, but instead specifying a certain quality metric like a certain percentage from the optimal objective. This would depend on the use-case and hence will not be considered here.

\subsection{\texorpdfstring{$k$}{k}-Nearest Neighbour (\texorpdfstring{$k$}{k}-NN) Loss}
An estimator of $\mathbb{E}_{c \sim \mathcal{C}_z}[c|z]$ that has lower variance than $c$ has to utilize the training data, while preferably being unbiased and consistent. This makes $k$-nearest neighbour ($k$-NN) regression a strong candidate as it is a non-parametric regression that has extensively studied strong asymptotic uniform consistency results \citep{Cheng1984} and more recently also finite-sample uniform consistency results \citep{Jiang2019}. Being non-parametric is beneficial as we will train a parameterized predictive model using signals from the estimator. If the estimator would be in the same model class, this would potentially amplify predictive and decision errors due to model misspecification. The $k$-NN estimator given feature values $z$ is defined as the arithmetic mean of $c_{(j)}$ where
\begin{align*}
    (z_{(j)}, c_{(j)}) = \argmin_{(z', c') \in D \backslash \bigcup_{i=1}^{j-1} \{(z_{(i)}, c_{(i)})\} }  ||z' - z|| 
\end{align*}
is the $j$-th closest data point in the feature space, $||\cdot||$ is some norm (we use the Euclidean norm) and $D$ the set of data points. The consistency of the $k$-NN estimator is intuitive, as with an increasing number of data realizations, the $k$ nearest neighbours converge in distance, i.e.,
\begin{align*}
    \lim_{|D| \rightarrow \infty} ||z_{(j)} - z|| < \epsilon \quad \forall j \in \{1, \dots, k\}, \forall \epsilon > 0.
\end{align*}

If the distance between feature values goes to zero, the conditional probability distributions $\mathcal{C}_{z_{(j)}}$ converge to $\mathcal{C}_z$ as well, assuming the underlying cumulative distribution function is continuous in $z$. This means that when the dataset grows infinitely large, the realized coefficient values $c_{(1)}, \dots, c_{(k)}$ are samples drawn from the same distribution $\mathcal{C}_z$. Taking the mean of this is then a sample mean and therefore a consistent estimator of $\mathbb{E}_{c \sim \mathcal{C}_z}[c|z]$ with variance $\text{Var}_{c \sim \mathcal{C}_z}(c) / k$.

These consistency results do not extend to the decision space, as optimization problem $x^*(\cdot)$ is a discontinuous multi-valued function. Despite this, we have that if the $k$ nearest neighbours are somewhat representative of a set of samples from conditional probability distributions $\mathcal{C}_z$, the obtained decisions $x^*(c_{(1)}), \dots, x^*(c_{(k)})$ could all be empirical optimal decisions given $z$. Since our $k$-NN estimator will still have some estimation error, we consider multiple potential optimal decisions in a similar fashion as in designing the top-$k$ loss. This way we naturally include the second principle we proposed for designing robust losses. The loss we propose is the average over the empirical regret of $k$ nearest neighbours:
\vspace{-1mm}
\begin{equation*}
    l_{k\text{-NN}}(\hat{c}, c) = \frac{1}{k}\sum_{j = 1}^k{c_{(j)}^w}^T  (x^*(\hat{c}) - x^*(c_{(j)}^w )),
\end{equation*}
where the values of the neighbours are adjusted based on some interpolation weight $w \in [0, 1]$:
\begin{align*}
     c_{(j)}^w = w c_{(j)} + (1-w) c. 
\end{align*}

We introduce this interpolation weight to make sure the $k$-NN estimator and therefore loss function remains distinct for all observations $(z, c)$. Without this weight, multiple different observations can have the same $k$ nearest neighbours and therefore the same estimator. This is not a general problem in $k$-NN regression, but it arises because we use the estimator in-sample, i.e., we provide an estimate of observations $c$ that have a known realized value. This problem arises especially when there is little data available as it leads to over-simplification of the existing relationships, i.e., increased assumption bias. This makes $w$ a tool to find a sweet spot in the bias-variance trade-off \citep{Geman1992}. 
In the same work, the $k$-NN estimator is considered as an example where increasing $k$ increases bias, while decreasing $k$ increases variance. Due to our in-sample usage of the $k$-NN estimator, varying $k$ does not solve the issue of different observations having equal estimators. We note that $w = 0$ makes the $k$-NN loss equal to the empirical regret loss and therefore this loss is more general.

\subsection{Applicability}
The second principle we introduce at the beginning of this section to design the proposed losses leads to adjusting target solutions, i.e., the solutions that are considered to be optimal given $c$. This only adjusts the second term of the empirical loss as shown in \eqref{eq:emploss}. It is important to note that gradients with respect to the predictive model parameters are independent of this second term. This means that both the RO loss and top-$k$ loss do not change the true gradient. However, this is also not our aim, as for the class of problems we look at \eqref{eq:sto} the true gradient is zero almost everywhere and therefore uninformative. In the experimental evaluation we apply our losses to two state-of-the-art gradient-approximation approaches, Smart `Predict, then Optimize' (SPO+) of \citet{Elmachtoub2022} and the Perturbed Fenchel--Young Loss (PFYL) of \citet{Berthet2020}. Similar to the approaches by \citet{Mulamba2021} and \citet{Sahoo2023}, these gradient approximation have a term consisting of the empirical optimal solution $x^*(c)$. Only the $k$-NN loss has the potential to generalize to approaches that are independent of empirically optimal solution $x^*(c)$. Technical details are provided in the Appendix.

\subsection{Computational Considerations}
Training a predictive model using gradient-based approaches requires frequently evaluating (solving) optimization problem $x^*(\cdot)$. In every epoch, $x^*(f_\theta(z))$ needs to be evaluated for each data point $(z,c)$ to obtain the loss and/or gradient. This is the most computationally expensive part of training and we will therefore quantify the computational expense as a number of problem evaluations. Looking at the empirical regret loss, $x^*(c)$ does not change during training, and can therefore be precomputed. Given $t$ data points and $s$ epochs, problem $x^*(\cdot)$ is solved $t*(s+1)$ times. 

The proposed robust losses increase precomputation time i.e., computation time before the training, but do not increase training time per epoch.  For the $k$-NN and top-$k$ loss the total number of evaluations is at most $t*(k + s)$. 
In
practice $s$ is significantly larger than $k$ and therefore the increase in total time is small. For the RO loss, the same number of evaluations is done compared to the empirical regret loss. While it could be that solving the robust problem formulation takes longer, this is also during precomputation at most $t$ times.


\section{Experimental Evaluation}
\label{sec:results}

We compare SPO+ and PFYL (number of samples $M$ = 1, perturbation amplitude $\sigma$ = 1) trained using empirical regret as loss to the same approaches using the presented robust losses (RO, top-$k$, $k$-NN), with PFL using mean squared error as an additional baseline. All predictive models are linear models. We use Python-based open-source package \termdefn{PyEPO} \citep{Tang2022} for the data generation of two experimental problems and the training, where the robust losses are implemented on top of the existing code. The $k$-NN loss is currently available in PyEPO. We use the Adam optimizer with learning rate 0.01 for the gradient descent and \termdefn{Gurobi} version 10.0.1 \citep{Gurobi2020} as the optimization problem solver.  We compare the results on three experimental problems: shortest path, travelling salesperson, and energy-cost aware scheduling. Additional details on the experimental problems are provided in the Appendix.  

For simplicity, we do not tune the hyperparameters of the robust losses, but use the same value for all problem configurations.  For the top-$k$ loss we use $k = 10$; the same for the $k$-NN loss where $w = 0.5$.  For the RO loss we set $\rho = 0.5$ and $\Gamma = \frac{n}{8}$, where $n = |c|$.  The batch size is 32.

\paragraph{Shortest path \& travelling salesperson.}   The shortest path problem considers a decision maker that has as goal to find the shortest path from start (NW) to end (SE) over a pre-specified grid ($10 \times 10$) with uncertain costs (objective coefficients), allowing only viable paths modelled by a vector of binary decision variables. The travelling salesperson problem also aims at finding a shortest path, but instead of a grid there is a set of fully connected nodes ($20$) that all need to be visited. For both problems data is generated as in PyEPO, with feature vectors $z$ of size 5 and the polynomial degree parameter fixed at 6. We look at different noise half-width values $\bar{\epsilon} \in \{0, 0.5, 1\}$ that multiply the generated objective coefficients by $\epsilon \sim \mathcal{U}(1 - \bar{\epsilon}, 1+\bar{\epsilon})$, and different training set sizes $t \in \{100, 1000\}$. We consider these parameter values to mimic different levels of aleatoric and epistemic uncertainty. The noise multiplies the existing data generating distribution by another distribution that increases variance and therefore aleatoric uncertainty, while the training set size is a measure of how much knowledge we have and therefore mimics epistemic uncertainty. In all cases a validation and test set of size 100 and 1000 are used respectively. The validation set is used to pick the best model found during training to prevent over-fitting, no model-selection is done.  200 epochs are run for t = 100, and 100 epochs for t = 1000. Since the data generation is a random process, we run 20 generated datasets to be able to measure significance.

\paragraph{Energy-cost aware scheduling.} As a third experimental problem we look at energy-cost aware scheduling \citep{Simonis1999} following precedent in a DFL setting \citep{Mandi2022}. The dataset consist of 789 days of historical 
energy price data at 30-minute intervals from 2011--2013 \citep{Ifrim2012}. The optimization problem consists of scheduling a given number of tasks ($3$) on a given number of machines ($10$) with a certain resource capacity. Each task has a given energy consumption per hour and the goal is to minimize the total daily energy consumption. Energy prices are uncertain and are therefore predicted. Instead of using the features in the dataset, we consider the predictive problem as a time series problem consisting of predicting the next day using the day before, both consisting of 48 values. This way the uncertain objective coefficients share the same feature vector as defined in problem formulation \eqref{eq:sto}. We also add noise to the data and we use training sizes of $t \in \{100, 500\}$ with a validation and test set of $100$ each. Given we have 789 days, we create 5 non-overlapping data partitions of the data preserving time order when $t = 100$. When $t = 500$ we use a single partition, and in both cases we omit the last 89 days. We run 50 epochs for every training run.

\begin{figure*}[tb]
    \centering
    \includegraphics{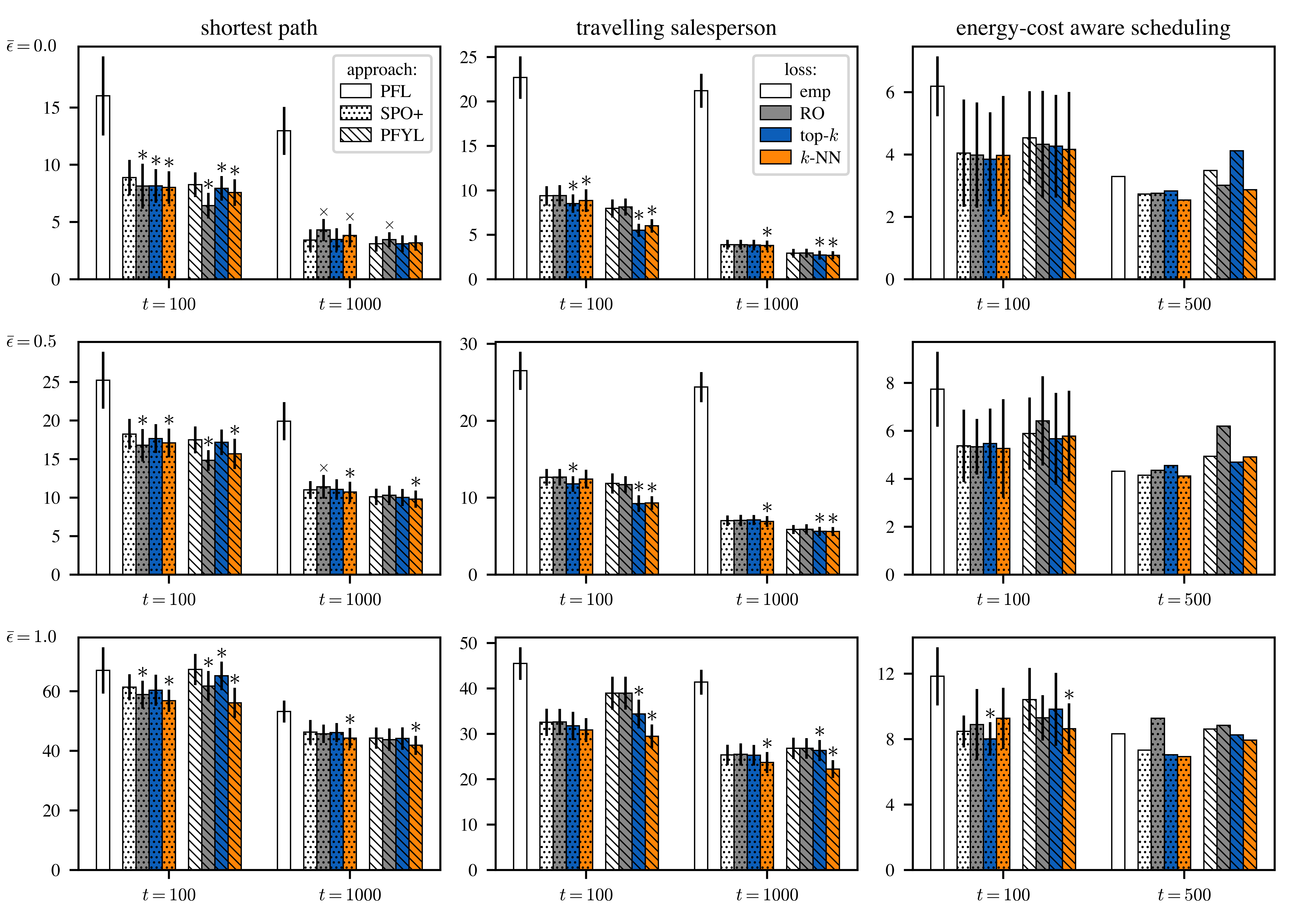}
    \vspace{-8mm}
    \caption{
    Test set mean normalized empirical regret in \% (y-axes) on 3 experimental problems with different noise ($\bar{\epsilon}$) and training size ($t$) values. Approaches are denoted by patterns; used losses by colour (mean squared error is used for PFL). Error bars denote one in-sample standard deviation on both sides of the mean. Mean values of robust losses that are significantly different (paired $t$-test, $\alpha = 0.05$) from their empirical regret counterpart are denoted with $*$ (better) or \small{$\times$} (worse). The results are shown as a table in the Appendix.}
    \label{fig:results}
\end{figure*}

\subsection{Results}
Figure \ref{fig:results} shows all results. For the shortest path and travelling salesperson, we see that when $t=100$ the robust loss performs significantly better than the empirical loss in 24 out of the 36 direct comparisons (equal problem, approach, $t$ and $\hat{\epsilon}$) and it never performs significantly worse. When $t=1000$, we see 13 significantly better and 4 significantly worse results, out of which 3 are on the shortest path problem with $\bar{\epsilon} = 0.0$. These results confirm our hypothesis that both more aleatoric and epistemic uncertainty lead to better performance of the robust losses. Due to the limited size of the dataset, we see only 2 significant results for the energy-cost aware scheduling problem. However, these are both in favour of the robust loss and in all other configurations a robust loss also performs best.

In general, the $k$-NN loss is the most effective loss, even when there is ample data. The RO loss is mostly effective for the shortest path, while the top-$k$ loss is more effective for the travelling salesperson. We hypothesize this is due to optimization problem characteristics, where the difference between optimal solutions and either robust optimal solutions (RO) or close-to optimal solutions (top-$k$) is most relevant.


\section{Discussion and Related Work}
\label{sec:rw}
Our proposed losses are evaluated on DFL gradient-approximation approaches \cited{Elmachtoub2022}{Berthet2020}, which assume a linear objective function. These approximations are needed since in general a quadratic objective is required to have defined non-zero gradients  \cited{Amos2017}{Veviurko2023}. Using the Karush-Kuhn-Tucker conditions, exact methods can be applied to augmented linear programs by adding a quadratic \citep{Wilder2019} or log-barrier \citep{Mandi2020} objective term. For MIPs, \citet{Ferber2020} introduce a cutting-plane approach to convert the discrete problem into an equivalent continuous problem, which is computationally expensive and therefore less applicable compared to gradient approaches. 

Gradient-approximation approaches have so far not been able to directly deal with a non-linear objective function or uncertain parameters in the constraints, but \citet{Silvestri2023} introduced score function gradient estimation for DFL, by predicting a distribution instead of a point. 

Empirical regret is in general used as a loss in DFL and therefore studied here, but some alternative losses have been proposed. \citet{Mulamba2021} introduce a contrastive loss that uses non-optimal decisions as negative samples to attain more effective gradients. \citet{Mandi2022} extend on this work by defining DFL as a learning-to-rank problem, i.e., accurately ranking multiple decisions leads to being able to pick an optimal decision and we also consider this to be more robust. Experimental results show that losses in these two works are especially effective in reducing training time for comparable decision quality. \citet{Shah2022} propose an approach to learn losses that are convex and therefore practical for training, however these losses are learned based on empirical regret. 

Since we show that the robust losses improve generalization when training data is limited, a natural alternative would be to apply regularization, as effectively used in deep learning \citep{Kukacka2017}. \citet{Elmachtoub2022} use a mean squared or absolute error term for regularization. This allows for improving prediction accuracy while preserving decision quality \citep{Tang2022}, but it does not improve decision quality which we do observe using robust losses.

\vspace{-0.1mm}
\section{Conclusion}
\label{sec:conc}

DFL is gaining considerable recent interest due to its effective end-to-end approach to data-driven optimization. This paper investigated the shortcomings of using empirical regret in DFL for MIPs, and proposed a trio of robust loss functions to mitigate the issues of poor generalization and biased learning. Experimental results show improved test--sample empirical regret, especially with little or noisy data, without inflating computational time. Future work includes studying the effectiveness of the robust losses relative to each other based on optimization problem characteristics, studying RO uncertainty sets and fine-tuning hyperparameters. A more general direction in DFL is analyzing robustness of DFL predictors in settings where at test time the data and/or the optimization problem are different, as this is currently not well-studied.

\section*{Acknowledgements}
We thank the reviewers for their suggestions. This work was partially supported by Epistemic AI and by TAILOR, both funded by 
EU Horizon 2020 under grants 964505 and~952215 respectively.

\bibliographystyle{named}
\bibliography{library}

\appendix
\section{Formal Proof Example 1}
In Section 2.2, we provide an example to show that DFL with empirical regret is biased towards high variance objective coefficients. This example is aimed at providing intuition and therefore does not contain a formal proof. Below we provide this proof. We note that this example is based on a simple optimization problem. Due to the complexity of combinatorial optimization problems in general, it is difficult to prove this bias in a general setting.

Recall the decision maker's optimization problem:
\begin{align} \tag{\ref{eq:sto}}
    \min_{x \in X} \mathbb{E}_{c \sim \mathcal{C}_z}[c^T x|z],   
\end{align}
where $X$ denotes the set of feasible decisions and $\mathcal{C}_z$ is the conditional distribution of uncertain objective coefficients $c \in \mathbb{R}^n$ given feature values $z \in \mathbb{R}^m$. Now we make the same assumptions as in the example, where (with abuse of notation) $c_i \in c, i \in \{1, \dots, n\}$ have some probability density function $c_i \sim f_i(x)$ and joint probability density function $c_i, c_j \sim f_{i,j}(x, y)$.

\begin{assumption} \label{ass:dec}
Decision set $X$ is finite and each decision has its own objective coefficient, i.e., $X = \{x : x \in \{0, 1\}^n, \sum_{i=1}^n x_i = 1 \}$.
\end{assumption}

\begin{assumption} \label{ass:sym}
    Assume that $c_i \in c, i \in \{1, \dots, n\}$ have a symmetrical probability distribution, i.e., $\forall i, \exists \bar{x}$: $f_i(\bar{x} + \delta) = f_i(\bar{x} - \delta)$ $\forall \delta \in \mathbb{R}$. 
\end{assumption}
\noindent In Section 2.2 we mention ``similarly shaped" distributions. For the proof we use the more specific Assumption \ref{ass:sym}.

\begin{assumption} \label{ass:ind}
    Assume that $c_i \in c, i \in \{1, \dots, n\}$ are independent, i.e., $f_{i, j}(x, y) = f_i(x)f_j(y),$ $\forall x, y$.
\end{assumption}

\begin{assumption} \label{ass:zero}
    Assume that $c_i \in c, i \in \{1, \dots, n\}$ have an expectation of zero, i.e., $\mathbb{E}_{c \sim \mathcal{C}_{z}}[c_i|z] = 0$, $\forall i$. (Note that under Assumption \ref{ass:sym} $\mathbb{E}_{c \sim \mathcal{C}_{z}}[c_i|z] = \bar{x}$.)
\end{assumption}
\noindent Assuming equal expectation is sufficient, however assuming an expectation of 0 makes the proof more readable. 

\begin{assumption}
    There exist $n_h$ decisions $X_h \subset X$ with coefficients with (high) variance $\Var(c^T x_h) = \sigma_h^2$ and $n_l$ decisions $X_l \subset X$ with coefficients with (low) variance $\Var(c^T x_l) = \sigma_l^2$, where $n_l + n_h = |X|$.
\end{assumption}
\noindent The distinction between low and high variance coefficients in combination with Assumption \ref{ass:sym} makes us able to relate certain probabilities and therefore introduce the following theorem:

\begin{theorem} \label{the:1}
     Assume $|X| > 2$. Given some $z$, we have that $\forall x_h \in X_h, x_l \in X_l$:
    \begin{align*}
    \mathbb{P}(x_h = x^*) > \mathbb{P}(x_l = x^*),
    \end{align*} when $\lim_{\sigma_l^2 \to 0}$, where $x^* = \argmin_{x \in X} \mathbb{E}_{c \sim \mathcal{C}_{z}}[c | z]$. 
\end{theorem}
\noindent We first sketch the proof in words, before providing the formal proof. When the low variance approaches zero, the probability that a decision with a low variance coefficient is optimal becomes small. Each other decision with a low variance coefficient has equal probability to be optimal, while the probability that all high variance coefficients are smaller than 0 is the product of multiple 50\% probabilities and therefore small. For the probability that a decision with high variance coefficient is optimal, similarly each other decision with a high variance coefficient has equal probability to be optimal, but it is likely that at least one of them is smaller than 0.

\begin{proof}
\begin{flalign*}
    &\lim_{\sigma^2_l \to 0} \mathbb{P}(x_l = x^* ) && \\
    &\stackrel{\text{a.s.}}{=}  \mathbb{P}(x_l = x^*; c^T x_l = \mathbb{E}_{c \sim \mathcal{C}_z}[c^T x_l | z], \forall x_l \in X_l) && \\
    &\stackrel{(\ref{ass:zero})}{=}  \mathbb{P}(x_l = x^*; \min_{x \in x} \mathbb{E}_{c \sim \mathcal{C}_{z}}[c^T x | z] = 0 ) && \\
    &= \mathbb{P}(x_l = x^*  | c^T x_h > 0, \forall x_h \in X_h ) \mathbb{P}(c^T x_h > 0, \forall x_h \in X_h) && \\
    &\stackrel{(\ref{ass:sym}, \ref{ass:ind})}{=} \frac{1}{n_l} (\frac{1}{2})^{n_h}, && \\
   & \lim_{\sigma^2_l \to 0} \mathbb{P}(x_h = x^* ) && \\
   &\stackrel{\text{a.s.}}{=}  \mathbb{P}(x_h = x^*; c^T x_l = \mathbb{E}_{c \sim \mathcal{C}_z}[c^T x_l | z], , \forall x_l \in X_l) && \\
    &\stackrel{(\ref{ass:zero})}{=}  \mathbb{P}( x_h = x^* ; \exists x_h \in X_h: c^T x_h > 0) \mathbb{P}( \exists x_h \in X_h: c^T x_h > 0)  && \\
    &\stackrel{(\ref{ass:sym}, \ref{ass:ind})}{=} \frac{1}{n_h} (1 - (\frac{1}{2})^{n_h}), &&
\end{flalign*}
where the numbers in brackets refer to the assumptions. When $n_l = n_h = 1$, both probabilities equal $\frac{1}{2}$, while increasing either $n_l$ or $n_h$ decreases $\mathbb{P}(x_l = x^* )$ more than $\mathbb{P}(x_h = x^* )$ which concludes the proof. 
\end{proof}

We note that Assumption~\ref{ass:dec} is quite restrictive and could be replaced by something more general:
\begin{assumption} \label{ass:new}
Decision set $X$ is finite and a one-to-one mapping $g: X \to Y$ exists such that we get the following alternative formulation of optimization problem \eqref{eq:sto}: $\min_{y \in Y} \mathbb{E}_{c_y \in \mathcal{C}_{y, z}}[c_y | z],$ where $c_y = c^T g^{-1}(y)$. 
\end{assumption}

The proof can be applied in the same way on this alternative formulation.  However note that Assumptions~\ref{ass:new} and~\ref{ass:ind} (independence) are contradictory.  This means the proof would not hold anymore, but intuitively, biased learning can still occur.

\newcommand{\sym}{\raisebox{3pt}{\tiny $\times$}}
\begin{table*}[tb]
\resizebox{\textwidth}{!}{
    \centering
    \begin{tabular}{l r l r r r r r r r r r}
    \toprule
        \multicolumn{3}{c}{~}&\multicolumn{1}{c}{PFL} & \multicolumn{4}{c}{SPO+} & \multicolumn{4}{c}{PFYL} \\ 
        \cmidrule(lr){4-4} \cmidrule(lr){5-8} \cmidrule(lr){9-12} 
        ~ & \multicolumn{1}{c}{$t$} & \multicolumn{1}{c}{$\bar{\epsilon}$} &\multicolumn{1}{c}{MSE} & \multicolumn{1}{c}{emp} & \multicolumn{1}{c}{RO} & \multicolumn{1}{c}{top-$k$} & \multicolumn{1}{c}{$k$-NN} & \multicolumn{1}{c}{emp} & \multicolumn{1}{c}{RO} & \multicolumn{1}{c}{top-$k$} & \multicolumn{1}{c}{$k$-NN}  \\ \midrule
        \multirow{6}{*}{\rotatebox[origin=c]{90}{SP}}& 100 & 0.0 & 16.0 (3.3) & 8.9 (1.4) & *8.1 (1.9) & *8.1 (1.4) & *8.0 (1.3) & 8.3 (1.0) & \textbf{*6.4 (1.0)} & *7.9 (1.0) & *7.6 (1.0) \\ 
         & 100 & 0.5 & 25.2 (3.5) & 18.2 (1.8) & *16.8 (1.9) & 17.7 (1.7) & *17.1 (1.7) & 17.5 (1.6) & \textbf{*14.8 (1.2)} & 17.2 (1.5) & *15.7 (1.8) \\ 
         & 100 & 1.0 & 67.0 (7.4) & 61.3 (3.9) & *58.9 (4.3) & 60.3 (4.8) & *56.8 (3.3) & 67.3 (4.8) & *61.7 (4.7) & *65.2 (4.4) & \textbf{*56.1 (4.6)} \\ 
         & 1000 & 0.0 & 13.0 (2.0) & 3.4 (0.8) & \sym4.3 (0.9) & 3.5 (0.9) & \sym3.8 (0.9) & \textbf{3.1 (0.6)} & \sym3.5 (0.5) & \textbf{3.1 (0.6)} & 3.2 (0.6) \\  
         & 1000 & 0.5 & 19.9 (2.3) & 11.0 (1.0) & \sym11.4 (1.4) & 11.1 (1.2) & *10.7 (1.2) & 10.1 (0.9) & 10.3 (1.1) & 10.0 (0.9) & \textbf{*9.8 (0.9)} \\ 
         & 1000 & 1.0 & 53.2 (3.3) & 46.3 (3.7) & 45.7 (2.6) & 46.1 (2.8) & *44.3 (3.0) & 44.3 (3.1) & 43.8 (3.3) & 44.1 (3.3) & \textbf{*41.9 (2.8)} \\ \midrule
        \multirow{6}{*}{\rotatebox[origin=c]{90}{TSP}} & 100 & 0.0 & 22.7 (2.2) & 9.4 (0.9) & 9.4 (1.0) & *8.5 (0.9) & *8.8 (1.1) & 8.0 (0.9) & 8.1 (0.8) & \textbf{*5.5 (0.6)} & *6.0 (0.6) \\ 
         & 100 & 0.5 & 26.5 (2.3) & 12.7 (0.9) & 12.7 (0.9) & *11.8 (0.9) & 12.4 (1.1) & 11.8 (1.1) & 11.7 (0.9) & \textbf{*9.2 (0.9)} & *9.3 (0.7) \\ 
         & 100 & 1.0 & 45.5 (3.3) & 32.5 (2.7) & 32.6 (2.6) & 31.7 (2.8) & 30.8 (2.4) & 38.9 (3.4) & 38.9 (3.3) & *34.4 (2.9) & \textbf{*29.4 (2.3)} \\ 
         & 1000 & 0.0 & 21.2 (1.8) & 3.9 (0.4) & 3.9 (0.4) & 3.8 (0.5) & *3.8 (0.4) & 2.9 (0.3) & 2.9 (0.4) & \textbf{*2.7 (0.4)} & \textbf{*2.7 (0.3)} \\ 
         & 1000 & 0.5 & 24.4 (1.8) & 7.0 (0.5) & 7.0 (0.6) & 7.1 (0.5) & *6.9 (0.5) & 5.9 (0.4) & 5.9 (0.5) & \textbf{*5.6 (0.4)} & \textbf{*5.6 (0.4)} \\ 
         & 1000 & 1.0 & 41.4 (2.5) & 25.3 (2.0) & 25.5 (2.2) & 25.3 (2.0) & *23.7 (2.0) & 26.8 (2.1) & 26.8 (2.0) & *26.3 (2.0) & \textbf{*22.2 (1.7)} \\ \midrule
         \multirow{6}{*}{\rotatebox[origin=c]{90}{ECAS}}
         & 100 & 0.0 & 6.2 (0.9) & 4.0 (1.7) & 4.0 (1.6) & \textbf{3.8 (1.5)} & 4.0 (1.9) & 4.5 (1.5) & 4.3 (1.7) & 4.3 (1.6) & 4.2 (1.8) \\ 
         & 100 & 0.5 & 7.7 (1.5) & 5.4 (1.5) & \textbf{5.3 (1.1)} & 5.5 (1.4) & \textbf{5.3 (2.0)} & 5.9 (1.5) & 6.4 (1.8) & 5.7 (1.9) & 5.8 (1.8) \\ 
         & 100 & 1.0 & 11.8 (1.7) & 8.5 (0.9) & 8.9 (2.1) & \textbf{*8.0 (0.9)} & 9.3 (1.8) & 10.4 (1.9) & 9.3 (1.3) & 9.8 (2.1) & *8.6 (1.5) \\ 
         & 500 & 0.0 & 3.3 \hspace{0.67cm} & 2.7 \hspace{0.67cm} & 2.8 \hspace{0.67cm} & 2.8 \hspace{0.67cm} & \textbf{2.5} \hspace{0.67cm} & 3.5 \hspace{0.67cm} & 3.0 \hspace{0.67cm} & 4.1 \hspace{0.67cm} & 2.9 \hspace{0.67cm} \\ 
         & 500 & 0.5 & 4.3 \hspace{0.67cm} & \textbf{4.1} \hspace{0.67cm} & 4.4 \hspace{0.67cm} & 4.5 \hspace{0.67cm} & \textbf{4.1} \hspace{0.67cm} & 4.9 \hspace{0.67cm} & 6.2 \hspace{0.67cm} & 4.7 \hspace{0.67cm} & 4.9 \hspace{0.67cm} \\ 
         & 500 & 1.0 & 8.3 \hspace{0.67cm} & 7.3 \hspace{0.67cm} & 9.3 \hspace{0.67cm} & 7.0 \hspace{0.67cm} & \textbf{6.9} \hspace{0.67cm} & 8.6 \hspace{0.67cm} & 8.8 \hspace{0.67cm} & 8.3 \hspace{0.67cm} & 7.9 \hspace{0.67cm} \\ \bottomrule
    \end{tabular}
    } \caption{Test set mean normalized empirical regret in \% (standard deviation) on 3 experimental problems with different noise ($\bar{\epsilon}$) and training size ($t$) values. The best performing approach and loss combination for each configuration is in \textbf{bold}. Mean values of robust losses that are significantly different (paired $t$-test, $\alpha = 0.05$) from their empirical regret counterpart are denoted with * (better) or $\sym$ (worse).
    Problems are abbreviated: shortest path (SP), traveling salesperson (TSP) and energy-cost aware scheduling (ECAS).}
    \label{tab:1}
\end{table*}

\section{Technical Details Applicability}
Section 3.4 describes the applicability of the robust losses. In this section we describe the gradients that are obtained when applying the robust losses on SPO+ and PFYL. Let us first introduce the (empirical regret based) SPO+ gradient \cite{Elmachtoub2022}
$$2(x^*(c) - x^*(2\hat{c} - c)),$$
and the PFYL gradient \cite{Berthet2020}
$$x^*(c) - \frac{1}{p} \sum_{i = 1}^p x^*(\hat{c} + \zeta_i),$$
where $\zeta_i$ are drawn from a distribution with a positive and differentiable density (we use $\mathcal{N}(0, \sigma = 1)$ in the experiments). Deriving the gradients for $l_\text{top-$k$}$ and $l_\text{RO}$ only requires substituting $x^*(c)$ by $\frac{1}{k} \sum_{j=1}^k x^*_{(j)}(c)$ and $x_{\text{RO}}^*(c)$ respectively. This works similarly for $l_\text{$k$-NN}$, where it is substituted by $\frac{1}{k}\sum_{j=1}^k x^*(c_{(j)}^w)$, except that in the case of the SPO+ gradient the second term is also adjusted to $\frac{1}{k}\sum_{j=1}^k x^*(2\hat{c} - c_{(j)}^w)$. To evaluate this gradient exactly the number of evaluations of $x^*(\cdot)$ per data point would be $k$ instead of $1$. Because of this we approximate this term with $x^*(2\hat{c} - \frac{1}{k}\sum_{j=1}^kc_{(j)}^w)$, which in some preliminary experiments showed to be as effective. 

\section{Experimental Problems Details}
This section contains further details on the experimental problem classes and the empirical results presented in Section 4.  First we provide some more details on how the experiments were run in general, followed by more details on the specific problems. 

As mentioned in the main article, the experiments were run using PyEPO \cite{Tang2022}. One main difference is that we included a validation set to pick the best model during training. Based on the results this did not change things significantly, however we believe this is best practice and easy to implement. It should positively impact methods that are prone to over-fitting, so given our claims this positively impacts the approaches trained with empirical losses compared to robust losses. 

The problems are all formulated as mixed-integer linear programs (MIPs) and solved using Gurobi \cite{Gurobi2020}. Below we specify the number features, the number of decision variables (and therefore objective coefficients) and sketch the MIP formulation for each problem. This gives some insight in the complexity of the predictive and optimization problem. 

\paragraph{Shortest path.} Given the pre-specified grid of $(v \times h)$, this problem has $v(h - 1) + h(v - 1)$ cost values and binary decision variables that make the objective, with constraints allowing only viable paths. To ensure this, constraints ensure that each grid location with an incoming arc also has an outgoing arc (except for the start an end points). This leads to $vh$ constraints.

\paragraph{Travelling salesperson.} Euclidean distance is used in between nodes, and since in an optimal decision each arc is only used in one direction, the problem is modeled as having $n(n-1)/2$ binary decision variables. Each location has exactly one incoming arc and one outgoing arc ($2n$ constraints). Additionally sub-tour elimination constraints ensure the solution is a single tour, which is based on the Dantzig-Fulkerson-Johnson formulation. These constraints can be added dynamically during solving, i.e., row generation, which is more efficient than specifying the exponentially many sub-tour constraints upfront. 

\paragraph{Energy-cost aware scheduling.} Predicting the following day using the previous day the predictive model has 48 features as input and 48 objective coefficients as output. The scheduling problem parameters are equal to the ones used in \textit{Energy-1} by \citet{Mandi2022}. This problem has constraints ensuring: Tasks start after their earliest start time, end before their latest end time and are only processed by a single machine. Also, each machines resource capacity per time unit is respected. Given $s$ tasks, $m$ machines and $t$ time, this leads to $3s + mt$ constraints. 


\section{Results Table}
Section 4.1 describes the experimental results, visualized by Figure~3. Table \ref{tab:1} presents the same results in a tabular format.

\end{document}